\documentclass[10pt,twocolumn,letterpaper]{article}
\usepackage{fullpage}

\usepackage[accsupp]{axessibility}
\usepackage{cvpr}      
\usepackage{float} 
\usepackage{graphicx}
\usepackage{amsmath}
\usepackage{amssymb}
\usepackage{booktabs}
\usepackage{xcolor}
\usepackage[symbol]{footmisc}
\usepackage{footnote}

\usepackage[pagebackref,breaklinks,colorlinks]{hyperref}

\usepackage[capitalize]{cleveref}
\crefname{section}{Sec.}{Secs.}
\Crefname{section}{Section}{Sections}
\Crefname{table}{Table}{Tables}
\crefname{table}{Tab.}{Tabs.}

\def\cvprPaperID{6451} 
\def\confName{CVPR}
\def\confYear{2024}

\begin{document}

\title{Relation Rectification in Diffusion Model}

\author{Yinwei Wu \quad \quad Xingyi Yang \quad \quad Xinchao Wang\footnotemark \\
National University of Singapore\\
{\tt\small wuyinwei@u.nus.edu, xyang@u.nus.edu, xinchao@nus.edu.sg}
}

\twocolumn[{%
\renewcommand\twocolumn[1][]{#1}%
\maketitle
\begin{center}
    \centering
    \captionsetup{type=figure}
   \includegraphics[width=0.95\linewidth]{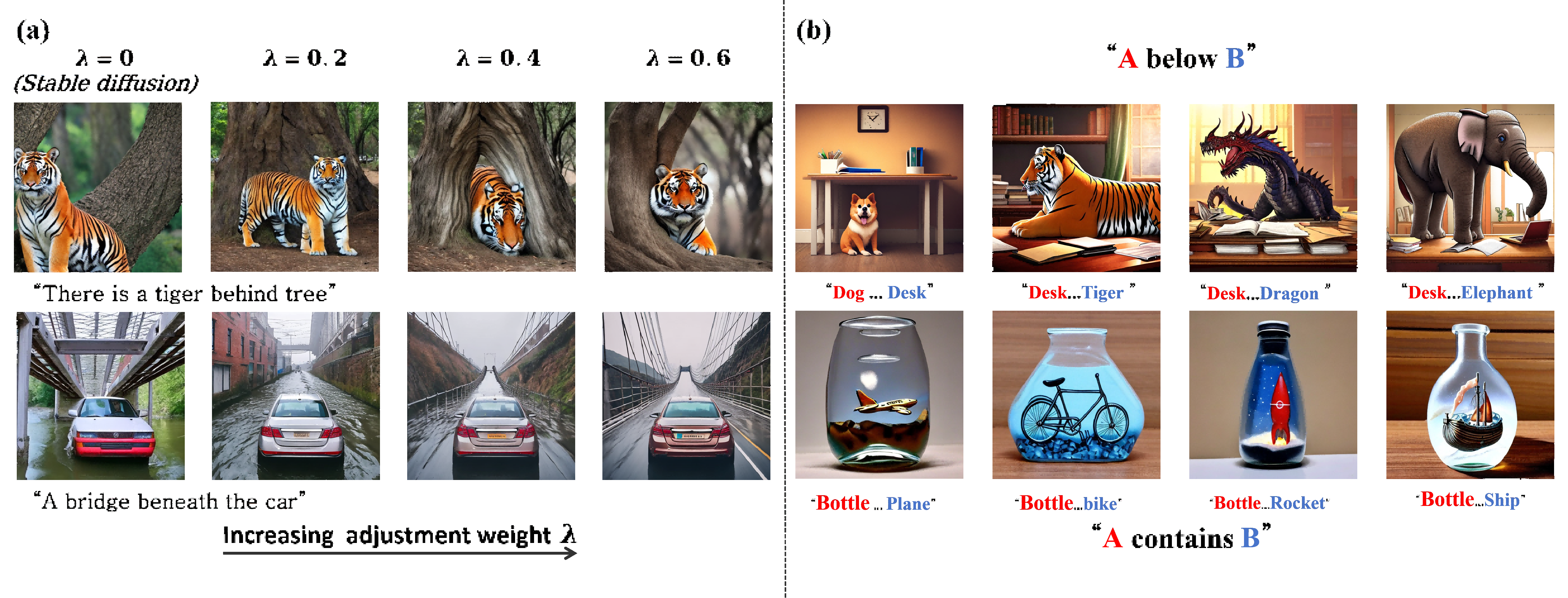}
   \caption{\textbf{Results Visualization of Relation Rectification.} (a) Our approach enables diffusion model to successfully generate images with the correct directional relation in response to the textual prompt, which they originally failed. (b) Our method can synthesize relation of diverse and unseen objects in zero-shot manner.}
   \label{fig:cover}
\end{center}%
}]


\footnotetext{* Corresponding Author}
\begin{abstract}
Despite their exceptional generative abilities, large T2I diffusion models, much like skilled but careless artists, often struggle with accurately depicting visual relationships between objects. This issue, as we uncover through careful analysis, arises from a misaligned text encoder that struggles to interpret specific relationships and differentiate the logical order of associated objects. To resolve this, we introduce a novel task termed \textbf{Relation Rectification}, aiming to refine the model to accurately represent a given relationship it initially fails to generate. To address this, we propose an innovative solution utilizing a Heterogeneous Graph Convolutional Network (HGCN). It models the directional relationships between relation terms and corresponding objects within the input prompts. Specifically, we optimize the HGCN on a pair of prompts with identical relational words but reversed object orders, supplemented by a few reference images. The lightweight HGCN adjusts the text embeddings generated by the text encoder, ensuring accurate reflection of the textual relation in the embedding space. Crucially, our method retains the parameters of the text encoder and diffusion model, preserving the model's robust performance on unrelated descriptions. We validated our approach on a newly curated dataset of diverse relational data, demonstrating both quantitative and qualitative enhancements in generating images with precise visual relations. Project page: \url{https://wuyinwei-hah.github.io/rrnet.github.io/} .
\end{abstract}

\section{Introduction}
\label{sec:intro}
Diffusion-based text-to-image (T2I) models\cite{Ho_Jain_Abbeel_Berkeley,Nichol_Dhariwal_2021} have set new benchmarks in the synthesis of images from textual descriptions, achieving remarkable fidelity and detail.
Nevertheless, a nontrivial gap persists in their ability to interpret and visually articulate the interacting objects within a given prompt, especially when the description includes directional or relational terms. For instance, a prompt such as ``\texttt{a book is placed on a bowl}'' frequently leads to a visual representation where the intended directionality of the interaction is misconstrued, resulting in scene akin to ``\texttt{a bowl is placed on a book}''. This indicates a crucial limitation in the model's relational understanding, a pivotal aspect of cognitive comprehension that remains to be fully integrated into the generated images. 

The problem of relation mis-interpretion is not unique to diffusion models; indeed, it is endemic to the broader class of Vision-Language Models (VLMs) \cite{Yuksekgonul_Bianchi_Kalluri_Jurafsky_Zou_2022}. The crux of this issue stems from the standard contrastive or mask-filling training of VLMs \cite{Thrush_Jiang_Bartolo_Singh_Williams_Kiela_Ross, Herzig_Mendelson_Karlinsky_Arbelle_Feris_Darrell_Globerson_2023, huang2023structure}, which prioritize global semantics but fails to capture correct relationships between objects. Consequently, these models, especially diffusion-based T2I models, often understand texts as ``Bag-of words''~\cite{Yuksekgonul_Bianchi_Kalluri_Jurafsky_Zou_2022}, neglecting the compositional semantics necessary for accurate image generation.

In response, some researchers have incorporated auxiliary input, such as canvas layouts, to guide image synthesis \cite{Zhang_Agrawala,Xie_Li_Huang_Liu_Zhang_Zheng_Shou_2023}. However, such interventions circumvent rather than resolve the primary limitation: the text encoder's inadequate response to the directionality of textual relations. Addressing this fundamental issue remains a pivotal concern for advancing the field.

To address the issue, we introduce a new task named \textbf{Relation Rectification}. Given a pair of prompts that describe the same relation but with the positions of the objects reversed (for instance, ``\texttt{The bowl is inside the cloth}'' vs. ``\texttt{The cloth is inside the bowl}''), termed object-swapped prompts (OSPs). The task aims to allow the model to give different responses to OSPs according to the differences in object relationships, rather than simplifying prompts to just "Bags-of-words".

Upon investigating the inner workings of the diffusion model, we found that the embedding of the special token $[EOT]$, signifying the \textit{end of text}, plays a pivotal role inguiding the generation of relationships. We further identified a critical issue: the embeddings of $[EOT]$ generated from OSPs are nearly identical, rendering the directionality of the relations indistinguishable.

To address this, we introduce \textbf{RRNet}, a novel framework designed to augment the relation understanding of diffusion models like Stable Diffusion (SD).  The essence of RRNet is to explicitly encode the directional aspect of relationships within a sentence. Specifically, we conceptualize OSPs as pairs of \emph{directed heterogeneous graphs}. To process these graphs, we utilize a Heterogeneous Graph Convolutional Network (HGCN), which generates adjustment vectors to distinctly separate the $[EOT]$ embeddings of OSPs. During training, we update the lightweight HGCN only, while maintaining the parameters of SD to be fixed. 

To evaluate the efficacy of our approach, we compiled \textbf{Relation Rectification Benchmark}, a new dataset for evaluation and rigorously test RRNet across a spectrum of relationships. The experimental results indicate that despite a minor decrease in image fidelity, RRNet enhances the accuracy of SD's relationship generation by up to 25
Additionally, our method significantly enhances interpretability, clearly depicting the directional transitions in relationships. Furthermore, RRNet demonstrates robust generalization capabilities, effectively handling even unseen objects in the dataset. This comprehensive testing underscores RRNet's potential in improving relationship interpretation in image generation tasks.


Our contributions are summarized as belows:
\begin{itemize}
    \item We introduce the novel task of \textbf{Relation Rectification}, focused on enhancing SD's capability to accurately generate images that reflect the directional relationships outlined in text prompts.
    \item We identify that the primary limitation of vanilla SD in relation rectification arises from the indistinguishable text embeddings of OSPs.
    \item We proposed \textbf{RRNet}, a HGCN based model that designed to aid SD in accurately generating images with directional relationships.
    Our approach requires only the training of a lightweight HGCN, and can effectively address the task of relation rectification.
    \item We contribute the \textbf{Relation Rectification Benchmark}, serving as a valuable evaluation tool for assessing the effectiveness of relation rectification methods.
\end{itemize}
\section{Related work}
\noindent\textbf{Diffusion Models.}
Diffusion models \cite{Nichol_Dhariwal_2021,Ho_Jain_Abbeel_Berkeley,Dhariwal_Nichol_2021,du2023stable,wang2024patch} view image generation as process of gradual denoising from isotropic noise. 
Recent advances have seen diffusion-based models reach the forefront in the field of T2I generation \cite{Gu_Chen_Bao_Wen_Zhang_Chen_Yuan_Guo_2022,Rombach_Blattmann_Lorenz_Esser_Ommer_2022,Saharia_Chan_Saxena_Li_Whang_Denton_Kamyar,Ramesh_Dhariwal_Nichol_Chu_Chen,Nichol_Dhariwal_Ramesh_Shyam_Mishkin_McGrew_Sutskever_Chen,yang2023diffusion}. Notably, the Stable Diffusion (SD) \cite{Rombach_Blattmann_Lorenz_Esser_Ommer_2022} operates by denoising within the latent space, conditioned on the text embeddings from pre-trained text encoders \cite{Radford_Kim_Hallacy_Ramesh_Goh_Agarwal_Sastry_Askell}. 
However, text embeddings generated from sentences that contain directional relations are often inaccurate, leading to difficulties for SD in accurately generating the relationships. Our method corrects these inaccurate embeddings. By doing so, it allows SD to more accurately capture relationship within text, leading to better image quality.

\noindent\textbf{Personalized Diffusion.}
Personalized diffusion focuses on creating images that align with specific, personalized visual concepts~\cite{Huang_Wu_Jiang_Chan_Liu_2023,Kumari_Zhang_Zhang_Shechtman_Zhu_2022,Gal_Alaluf_Atzmon_Patashnik_Bermano_Chechik_Cohen,Ruiz_Li_Jampani_Pritch_Rubinstein_Aberman_2022} by tuning the general-purpose diffusion model. Leveraging off-the-shelf model tuning techniques~\cite{Hu_Shen_Wallis_Allen-Zhu_Li_Wang_Chen_2021,yang2022deep,houlsby2019parameter}, it modifies aspects such as text embeddings or attention patterns within the diffusion process. These modifications guide the diffusion to achieve a customized image generation. 
Unlike other methods that concentrate on generating particular visual content, we focus on improving the model's ability to generate precise visual relationships by adjusting the embeddings.

\noindent\textbf{Vision-Language Models.} Vision-Language Models (VLMs) \cite{Radford_Kim_Hallacy_Ramesh_Goh_Agarwal_Sastry_Askell, Li_Li_Xiong_Hoi, Jia_Yang_Xia_Chen_Parekh_Pham_Le_Sung_Li_Duerig_2021} aim to learn a unified cross-modality representation space between image and text by pre-training on a large number of image-text pairs.
For example, CLIP \cite{Radford_Kim_Hallacy_Ramesh_Goh_Agarwal_Sastry_Askell} learned text-image shared representations by using contrast learning on large-scale image-text pairs. However, recent work~\cite{Yuksekgonul_Bianchi_Kalluri_Jurafsky_Zou_2022} suggests that the contrastive objective used to train CLIP does not explicitly encourage it to learn sentence order information, leading to CLIP's tendency to interpret sentences as Bags-of-Words. Our work not only identifies the impact of this characteristic on T2I diffusion models which using CLIP as a text encoder, but also proposes an effective solution to mitigate the negative effects it introduces.

{\noindent\textbf{Compositional Image Generation.} 
Compositional Image Generation aims to enable generative models to generate images from prompts that describe the composition of certain concepts.
Composable Diffusion \cite{Liu_Li_Du_Torralba_Tenenbaum_2022} utilizes multiple diffusion models to control the generation of different concepts, thereby achieving precise object positioning.
GLiGEN \cite{Li_Liu_Wu_Mu_Yang_Gao_Li_Lee_2023} and ControlNet \cite{Zhang_Agrawala} guide compositional generation by providing additional multimodal conditions for training.
Another branch of works \cite{Xie_Li_Huang_Liu_Zhang_Zheng_Shou_2023,Feng_He_Fu,Chen_Laina_Vedaldi_2023,Chefer_Alaluf_Vinker_Wolf_Cohen-Or_2023} modify the cross-attention layout within the diffusion model to control object generation positions in a training-free manner.
However, previous research primarily examines simple~\cite{Gokhale_Palangi_Nushi_Vineet_Horvitz_Kamar_Baral_Yang} spatial relationships. Our work delves into more intricate spatial and action-based relationships.
}

\noindent\textbf{Graph Convolutional Network.}
Graphs differ fundamentally from image and text data. They comprise nodes and edges, with edges representing the relationships between nodes. Recent advancements in Graph Convolutional Networks (GCNs) \cite{Kipf_Welling_2016, 
 Hamilton_Ying_Leskovec_2017, Liu_Zhou_2020} have significantly improved graph learning tasks. GCNs have been successfully applied in diverse domains \cite{Rezvanian_Meybodi_2016, Li_Yuan_Radfar_Marendy_Ni_O, Jing_Mao_Yang_Zhan_Song_Wang_Tao_2022, Wu_Chen_Shen_Guo_Gao_Li_Pei_Long_2023}.
Heterogeneous graphs represent a distinct category of graphs, and they are capable of depicting various types of nodes and their interrelations. To manage this particular type of graph, Heterogeneous GCNs (HGCN) \cite{Wang_Ji_Shi_Wang_Ye_Cui_Yu_2019, Zhang_Song_Huang_Swami_Chawla_2019} are proposed. More specifically, HGCNs adopt various functions to handle information emanating from various types of nodes.
In our work, we introduce HGCN to tackle the challenge posed by the text encoder of SD, specifically its insensitivity to the directional relations within sentences.



\section{Approach}\label{approach}



\subsection{Preliminaries}

\begin{figure}[t]
  \centering
   \includegraphics[width=1\linewidth]{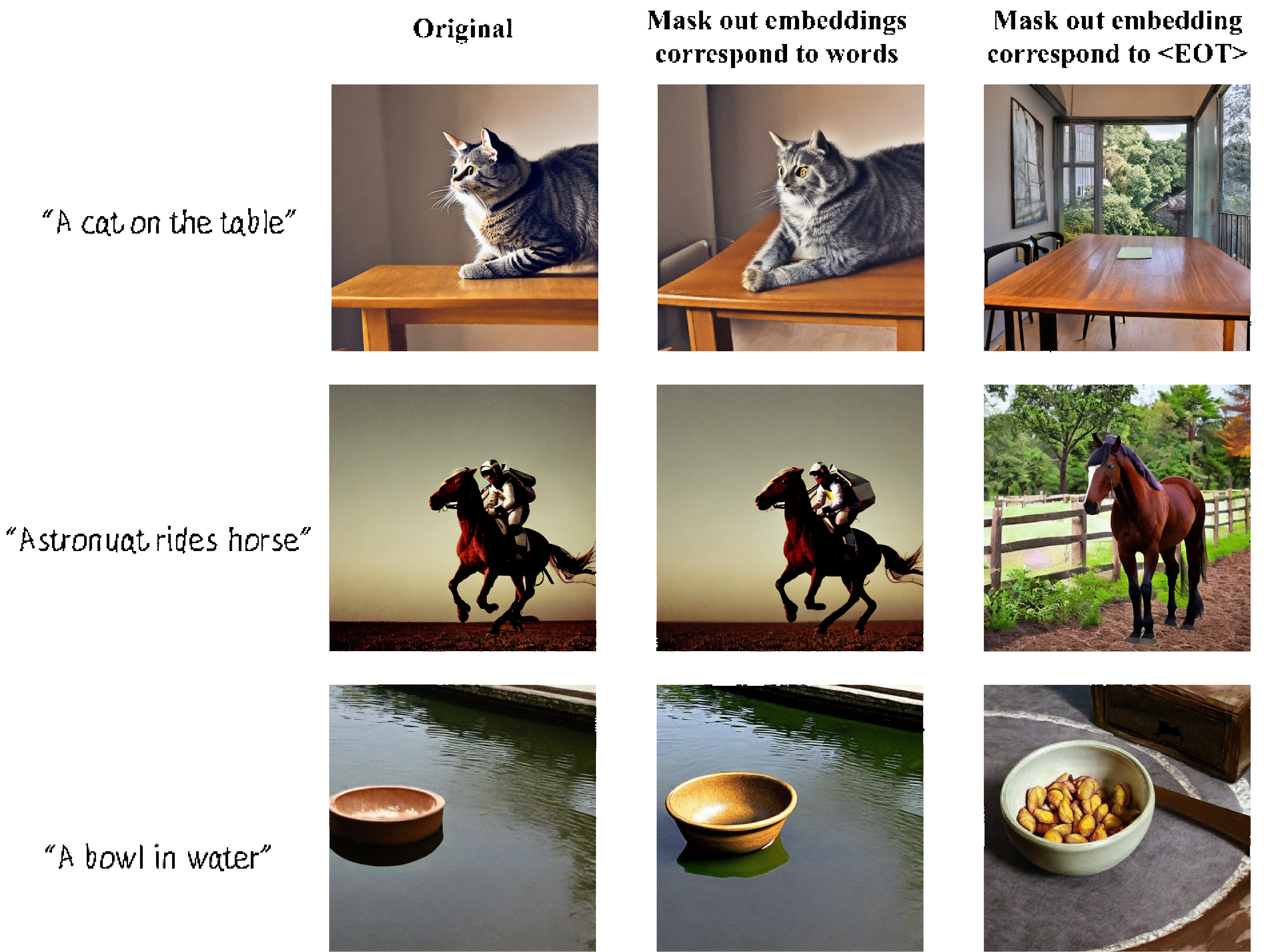}
   \caption{\textbf{Effect of masking out text embeddings corresponding to different tokens.} We found that masking out the embedding of $[EOT]$ dramatically destroy the semantic of generated images, including relationships, whereas masking out the embeddings corresponds to words  results in only marginal changes.}
   \label{fig:mask}
\end{figure}
\noindent\textbf{Text-to-Image Diffusion Models and its Text Encoder}.
We apply our method over a pretrained T2I SD.  
Given an initial Gaussian noise $z \sim \mathcal N(0, I)$ and a textual prompt $y$, the SD can generate an image that matches the prompt by stepwise denoising. The SD is trained by minimizing LDM loss as follows:
\begin{equation}
  \mathcal L_{LDM} = \mathbb{E}_{z \sim \mathcal N(0, I), y, t}[||\epsilon-\epsilon_{\theta}(z_t, t, c(y))||^2_2], 
  \label{eq:LDM_loss}
\end{equation}
where the $\epsilon_{\theta}$ is parameterized as a U-Net. At each timestep $t$, the denoising network $\epsilon_{\theta}$ aims at predicting the noise $\epsilon$ added to the latent code. The textual prompt $y$ that describes the final image content guides the denoising process.
Before provided to the SD, the prompt is encoded by a text encoder $c$, to produce text embeddings $c(y) \in \mathbb{R}^{K \times d}$, where $K$ represents the number of tokens and $d$ the token embedding dimension. Typically, the text embeddings $c(y)$ generated by $c$ have a length $K$ of 77.
 In addition to embeddings representing each word in $y$, embeddings of special tokens are also included. Notably, the special token $[EOT]$ marks the \textit{end of Text}, and its embedding (denoted as $V_{eot}$) is used to represent semantic information for the entire sentence during the contrast learning of $c$.  


\noindent\textbf{Key Finding: $V_{eot}$ controls the relation.} In addressing our specific problem, we identified that the $V_{eot}$ vector plays a pivotal role in controlling the relationships and semantics in generated images. 
Our experiments, as shown in Figure \ref{fig:mask}, revealed a critical observation: when $V_{eot}$ is masked out, SD struggles to generate images that accurately depict valid relationships. This finding underscores the heavy reliance of SD on $V_{eot}$ for relation generation. We hypothesize that this is due to $V_{eot}$ accumulating rich semantics from other tokens, including both object and relational information, thereby serving as a crucial component in relationship generation. Further, we observed that the $V_{eot}$ vectors of OSPs have a cosine similarity close to 1, indicating they are nearly indistinguishable to SD. 

Based on these findings, we plan to differentiate the $V_{eot}$ vectors of OSPs to enhance SD's relationship generation accuracy.

\noindent\textbf{Graph Convolutional Network.} GCN excels in processing graph data, A graph can be defined as a tuple of $G=(\mathcal{V}, \mathcal{E})$, $\mathcal{V}$ represents for nodes set and $\mathcal{E}$ is the edges set. For the directed graph used in our work, each directed edge $e_{i,j} \in \mathcal{E}$ connects node $v_i$ to node $v_j$.

GCN updates a node's representation by aggregating information from its neighbors.  For a GCN comprising $L$ convolutional layers, the update formula for the node representation of the $l$ th layer can be expressed as follows:
\begin{equation}
   h_i^{(l+1)}=\sigma(b^{(l)}+\sum_{j:(e_{j,i}) \in \mathcal{E}}\alpha_{j,i}h_j^{(l)}W^{(l)}),
  \label{eq:graph_conv}
\end{equation}
where $h^{(l)}$ is the representation of the node at layer $l$. $\alpha_{j, i}$ denotes the edge weight. $W$ and $b$ are learnable parameters and $\sigma$ is the activation function.

To differentiate the $V_{eot}$ of OSPs, we use graphs to represent the directions of the relations in sentences. Our approach utilizes a heterogeneous graph to model diverse information as different node types. Objects and relations are distinct node types, $\mathcal{V}_O, \mathcal{V}_R \in \mathcal{V}$. Their information will be aggregated into $\mathcal{V}_{\Delta EOT}$, which is responsible for learning adjustment vectors to separate the $V_{eot}$ of OSPs. 

In our HGCN, the weight matrix $W$ varies by node types, allowing specific aggregation for each information type.

 \subsection{Problem Definition}

 


\begin{figure*}[t]
  \centering
   \includegraphics[width=0.9\linewidth]{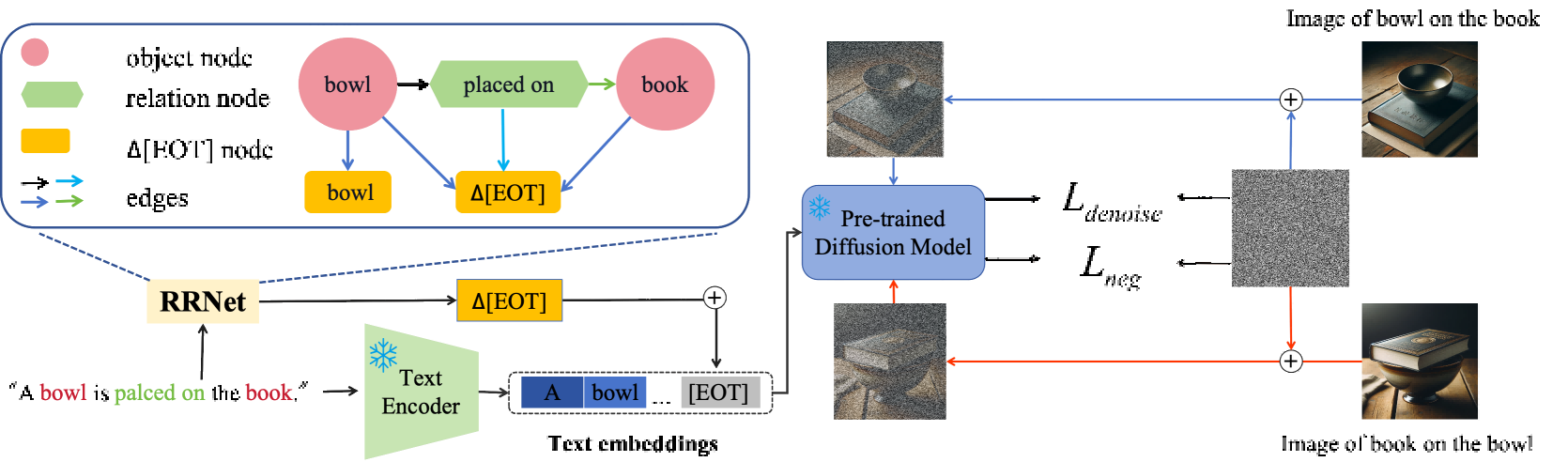}

   \caption{\textbf{Our RRNet Architecture.}
   Given OSPs and their exemplar images, the RRNet learns to produce adjustment vectors $h_{\Delta EOT}^{(L)}$, which will then be added on original $V_{eot}$ of the prompts. The rectified embeddings then will used as the condition to guidance the generation process of a frozen SD. The upper left part is the heterogeneous graph RRNet uses to model the relation direction. Upon optimization with negative loss and denoising loss, the SD will be able to generate images with correct relation direction.}
   \label{fig:framework}
\end{figure*}




 In our paper, we define \emph{relation rectification}  as the task of enabling T2I SD to generate images that more accurately represent the described relationships of OSPs. For a prompts $y$ contains directional relationship, 
 we aim to adjust its text embeddings $c(y)$ using an additional model $\phi$ such that:
 \begin{equation}
   \arg\max_{\phi} \, P(x|\phi(c(y))) - P(\widetilde{x}|\phi(c(y)))
  \label{eq:problem difinition1}
\end{equation}
where $P$ is the generating distribution of SD, and $x$ and $\widetilde{x}$ are two sets of images describe by $y$ and its object-swapped counterpart $\widetilde{y}$. 
Achieving this objective requires two key criteria:

\noindent\textbf{C1.} The text embeddings of OSPs must be distinguishable after adjustment.

\noindent\textbf{C2.} The adjusted embeddings should remain valid for SD’s input domain, effectively guiding the generation of the correct relational direction.

As such, we give our solution to each criterion.

\noindent\textbf{Solution to C1.} We introduce the Relation Rectification Net (RRNet), which conceptualizes OSPs as heterogeneous graphs to capture the directional relations. RRNet uses these graphs to produce adjustment vectors that separate the embeddings of OSPs. We discuss RRNet in Sec \ref{Text Embedding Separation}.

\noindent\textbf{Solution to C2.} We utilize SD's capability to understand visual semantics, guiding RRNet to create effective adjustment vectors. By analyzing example images corresponding to OSPs, SD discerns various relational directions, imparting this knowledge to RRNet for modifying text embeddings. More details are discussed in Sec \ref{two_loss}.







\subsubsection{Relation Rectification Net}\label{Text Embedding Separation}
As we have identified, the relationship information for SD is predominantly encoded within the $V_{eot}$. To effectively rectify and optimize this vector for accurate relation generation, we've developed a model that incorporates a HGCN. This HGCN is specifically engineered to process heterogeneous graphs derived from OSPs, subsequently outputting adjustment vectors that refine the $V_{eot}$.
\noindent\textbf{Heterogeneous Graph Construction}.
A pair of OSPs contain objects $A, B$ and a relationship $R$ conceptualized as two distinct triplets: $<A, R, B>$ and $<B, R, A>$. It is important to note that $<A, R, B>\neq<B, R, A>$   due to the directional nature of $R$. These triplets are directly modeled as directed graphs, for example $<A, R, B>$ as a graph with directed edges $A \rightarrow R$ and $R \rightarrow B$.

Considering the different semantics of object and relation nodes, we employ heterogeneous graphs. Here, objects and relations are represented as two distinct node types, $v_r \in \mathcal{V}_R$ for relations and and $v_o \in \mathcal{V}_O$ for objects.

In addition to $\mathcal{V}_R$ and $\mathcal{V}_O$, a third kind of nodes, $\mathcal{V}_{\Delta EOT}$, is utilized to learn the adjustment vector $h_{\Delta EOT}$. The learned $h_{\Delta EOT}$ will be used to adjust the relation direction information in original $V_{eot}$ of OSPs. 




\noindent \textbf{Relation Adjustment.} With the graph topology established, we initiate the node representations.  For object nodes $\mathcal{V}_O$ and relation nodes $\mathcal{V}_R$, we employ CLIP's word embeddings, rich in semantic content, as their initial representations. Node $\mathcal{V}_{\Delta EOT}$ is initialized randomly.

In the HGCN layers, the information from $\mathcal{V}_O$ and $\mathcal{V}_R$ aggregates into $\mathcal{V}_{\Delta EOT}$ along edges. The update formula for $\mathcal{V}_{\Delta EOT}$ is:
\begin{equation}
   h_{\Delta EOT}^{(l+1)} = \sum_{\mathcal{E} \in \hat{\mathcal{\mathcal{E}}}}\sigma(b^{(l)}+\sum_{j:(e_{j,i}) \in \mathcal{E}}\alpha_{j,i}h_j^{(l)}W_{\mathcal{E}}^{(l)}), 
  \label{eq:gcn_eot}
\end{equation}
where $\hat{\mathcal{E}}$ is the set of all edge types, and $W_{\mathcal{E}}^{(l)}$ denotes the weight of the edge of type $\mathcal{E}$. 
After several convolution layers, relation direction and node representations of $A, R$ and $B$ merge into $\mathcal{V}_{\Delta EOT}$. As illustrated in the bottom left part of Figure \ref{fig:framework}, the final adjusted $\overset{*}{V_{eot}}$ is obtain via:

\begin{equation}
   \overset{*}{V_{eot}} = V_{eot} + \lambda \cdot h_{\Delta EOT}^{(L)}, 
  \label{eq:adjust}
\end{equation}
here the $\lambda \in [0, 1]$ moderates the adjustment strength.
By adjusting the value of $\lambda$, a trade-off between image quality and relation generation accuracy can be made.
This additive operation is inspired by previous works \cite{Chuang_Jampani_Li_Torralba_Jegelka_2023,Trager_Perera_Zancato_Achille_Bhatia_Soatto_2023} on concept algebraic in VLMs' embedding space. 




\noindent \textbf{Object Node Disentanglement.} 
Our objective extends beyond just memorizing relationships; we aim to separate the features of individual objects, represented by nodes $A$ and $B$, from the relationship $R$. This separation is essential to prevent the blending of object features with relations.

To achieve this, as depicted in Figure \ref{fig:framework}, RRNet includes a dedicated node, $v_{\Delta EOT}^A \in \mathcal{V}_{\Delta EOT}$ for object $A$. $V_{eot}^A$ is extracted from template sentence ``\texttt{This is a photo of \{A\}}''. We aim to preserve $A$ 's accurate semantic in $V_{eot}^A$ after applying the adjustments through equation \ref{eq:adjust} with $h_{\Delta EOT}^A$. The $h_{\Delta EOT}^A$ is obtained from the final layer representation of $v_{\Delta EOT}^A$, as depicted in equation \ref{eq:gcn_eot}. This method ensures the disentanglement of node $A$ from node $B$ and relation $R$. 

However, since $V_{eot}^A$ already aligns closely with object $A$'s semantic and might not require adjustment, $h_{\Delta EOT}^A$ could potentially learn a trivial solution, such as zero. To prevent this, we add Gaussian noise to $V_{eot}^A$, so that the $h_{\Delta EOT}^A$ is forced to learning a valid semantic vector.

For the disentanglement of object B, owing to the concurrent training of RRNet from $<A, R, B>$ and $<B, R, A>$ directions, when training in the $<B, R, A>$ direction, B is also disentangled
\subsubsection{Relation Compelling Losses}\label{two_loss}

In our study, we have developed a relation compelling loss, consisting of both \emph{positive} and \emph{negative} components, tailored to enhance SD's interpretation of relationships.

\noindent\textbf{Positive Loss.} Our goal is to guide SD in identifying text embeddings that generate images with specific relationship semantics, as outlined in equation \ref{eq:problem difinition1}.  To achieve this, we directly utilize the denoising loss as below:
\begin{equation}
\begin{split}
  \mathcal L_{denoise} = \mathbb{E}_{z \sim \mathcal N(0, I), y, t}[||\epsilon-\epsilon_{\theta}(x_t, t, \phi(c(y)))||^2_2],
\end{split}
  \label{eq:n_denoise_loss}
\end{equation}
where $\phi$ is the RRNet, and $\phi(c(y)))$ adjusts the text embeddings of prompt $y$. The $x_t$ are exemplar images corresponding to $y$. Through $\mathcal L_{denoise}$, RRNet learns to generate adjustment vectors that align text embeddings of $y$ with the relation semantics in the exemplar images. 


\noindent\textbf{Negative Loss.} Solely relying on $\mathcal L_{denoise}$ risks the model focusing on superficial features such as image background, rather than the desired relations. To address this and ensure $V_{eot}$ separation in OSPs, we introduce a negative loss: 
\begin{equation}
\begin{split}
  \mathcal L_{neg} = \mathbb{E}_{z \sim \mathcal N(0, I), \widetilde{y}, t}[-||\epsilon-\epsilon_{\theta}(x_t, t, \phi(c(\widetilde{y})))||^2_2],
\end{split}
  \label{eq:n_loss}
\end{equation}
where $\widetilde{y}$ is the OSP counterpart of $y$. 

The final loss becomes:
\begin{equation}
\begin{split}
  \mathcal L = \eta \cdot  \mathcal L_{denoise} + \xi \cdot  \mathcal L_{neg}
\end{split}
  \label{eq:final_loss}
\end{equation}
here, the $\eta$ and $\xi$ ere hyperparameters balancing the two losses. 
Intuitively, $\mathcal L_{denoise}$ helps RRNet in discerning relation semantics, while $L_{neg}$ mitigates unintended semantics, aiding in separating $V_{eot}$ of OSPs.

During training, for each OSP pair represented as $<A, R, B>$ and $<B, R, A>$, our dataset includes four types of image-text pairs: two types for the original OSPs and two for disentangling the object node features. The computation of loss for each image-text pair involves the incorporation of the other three pairs using Equation \ref{eq:final_loss}.

Once trained, RRNet is capable of generating adjustment vectors that effectively separate the $V_{eot}$ of OSPs, thus enhancing SD's precision in relationship generation.

\section{Experiment}
\subsection{Experimental Setup}

\noindent\textbf{Dataset.} For a comprehensive benchmarking, we compiled a dataset with 21 relationships, split into 8 positional (like \texttt{below}, \texttt{on the left}) and 13 action (such as \texttt{touch}, \texttt{follow}) types. Each includes a pair of object-swapped prompts (OSPs) and corresponding images. We generated 100 images per prompt, totaling 4200 images, to rigorously assess the accuracy of relation generation. For detailed dataset statistics, please refer to Appendix \ref{ES}.


\begin{figure}[!htb]
  \centering
   \includegraphics[width=0.9\linewidth]{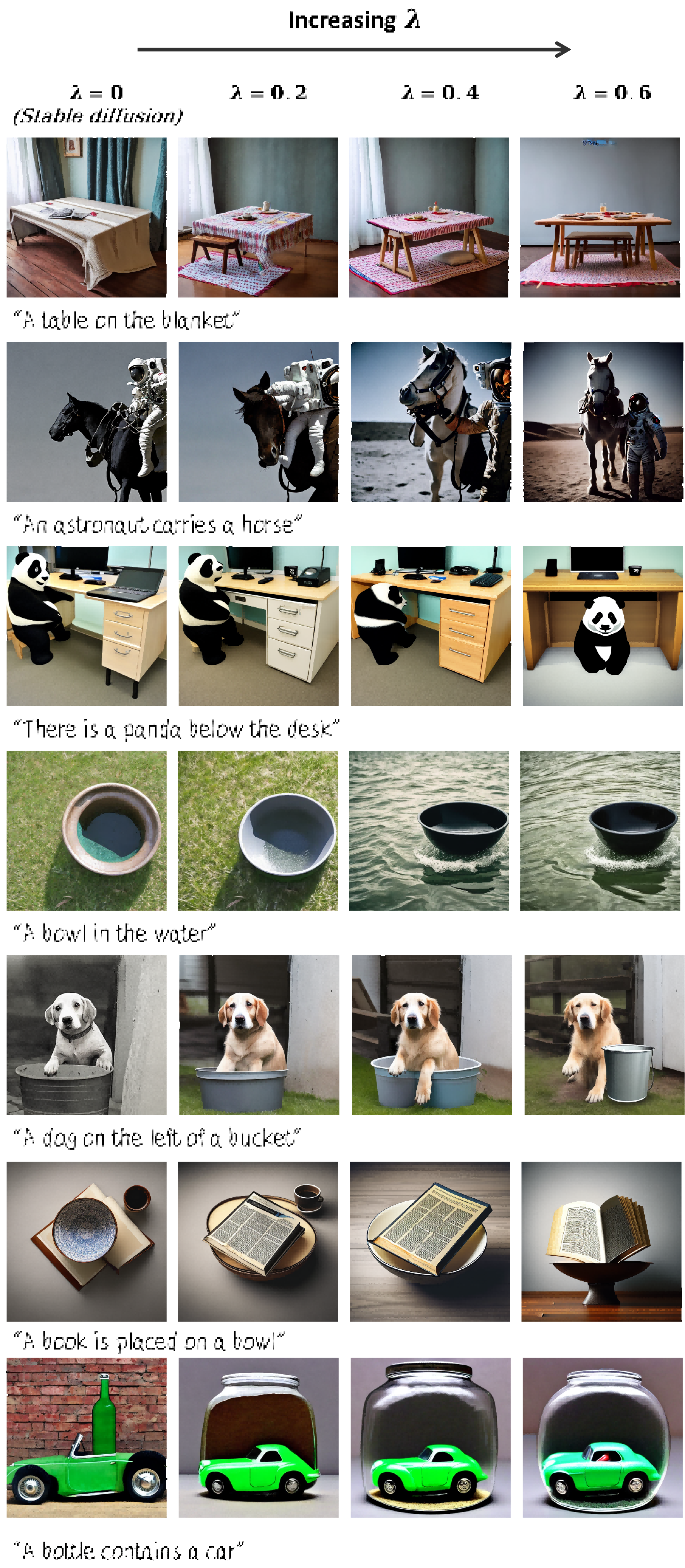}

   \caption{\textbf{Qualitative Results.} By increasing the weight $\lambda$ of the adjustment vector, we show the process of correcting relation direction in the generated images that were originally incorrect.}
   \label{fig:change}
\end{figure}

\noindent\textbf{Implementation Details}.
We train RRNet on Stable Diffusion 2-1 for 100 epochs. During training, $\lambda$ in Equation~\ref{eq:adjust} is set to 1. For the loss in Equation~\ref{eq:final_loss}, we found that setting the weight $\eta$ of denoising loss to 10 and weight $\xi$ for negative loss to 2 works well for most cases. Training for each relationship only takes about 20 minutes on 1 NVIDIA V100 GPU.

During generation, the weight $\lambda$ can be set within $[0,1]$. In our experiments, we assess the generation outcomes using various values of $\lambda$.
For the denoising process, we use the PNDM scheduler \cite{karras2022elucidating} and adopt 30 steps. The classifier-free guidance is applied for better image quality.


\noindent\textbf{Baselines.}  In the absence of existing approaches designed to generate correct relations, we established our own baselines. The first baseline is SD itself. Additionally, we compare our method with personalized diffusion models~\cite{Kumari_Zhang_Zhang_Shechtman_Zhu_2022, 
Gal_Alaluf_Atzmon_Patashnik_Bermano_Chechik_Cohen, Ruiz_Li_Jampani_Pritch_Rubinstein_Aberman_2022}, where we optimize the CLIP text encoder within the SD model. To ensure a fair comparison, we optimize the text encoder using the same loss function that is applied in these personalized diffusion methods. More comparative experiments are provided in the Appendix \ref{AC}.

\noindent\textbf{Evaluation Metrics.} We report two evaluation metrics

\begin{itemize}
    \item \textbf{Relationship Generation Accuracy}. To evaluate the accuracy of relationship generation, we employ vision-language chatbots proficient in image semantics, specifically Qwen-VL-Chat \cite{Bai_Bai_Yang} and LLaVA \cite{Liu_Li_Wu_Lee}.
    The evaluation approach involves:
    (1) Using a sentence representing triplet \( <A, R, B> \) and a corresponding image as a prompt. (2) Verifying the presence of entities A and B in the image. (3) Guiding the chatbots to determine the relationship between A and B. (4) Having the chatbots choose the most plausible relationship from \( <A, R, B> \), \( <B, R, A> \), or \( Neither \).
    
    In steps (1) and (2) described above, chatbots perform the task of detecting object generation, from which we derive the Object Generation Accuracy (OGA). This metric is used to assess the accuracy of entity generation.

    \item \textbf{Fréchet Inception Distance (FID)}. FID \cite{Heusel_Ramsauer_Unterthiner_Nessler_Hochreiter_2017} is used to evaluate the quality of generated images. It can measure the feature similarity between generated and real images via a pre-trained inception model \cite{Szegedy_Vanhoucke_Ioffe_Shlens_Wojna_2016}. We caculate the FID score between 4,200 generated images and all images from validation set of COCO \cite{Lin_Maire_Belongie_Hays_Perona_Ramanan_Dollár_Zitnick_2014}.

\end{itemize}
\begin{figure}[!h]
  \centering
   \includegraphics[width=0.9\linewidth]{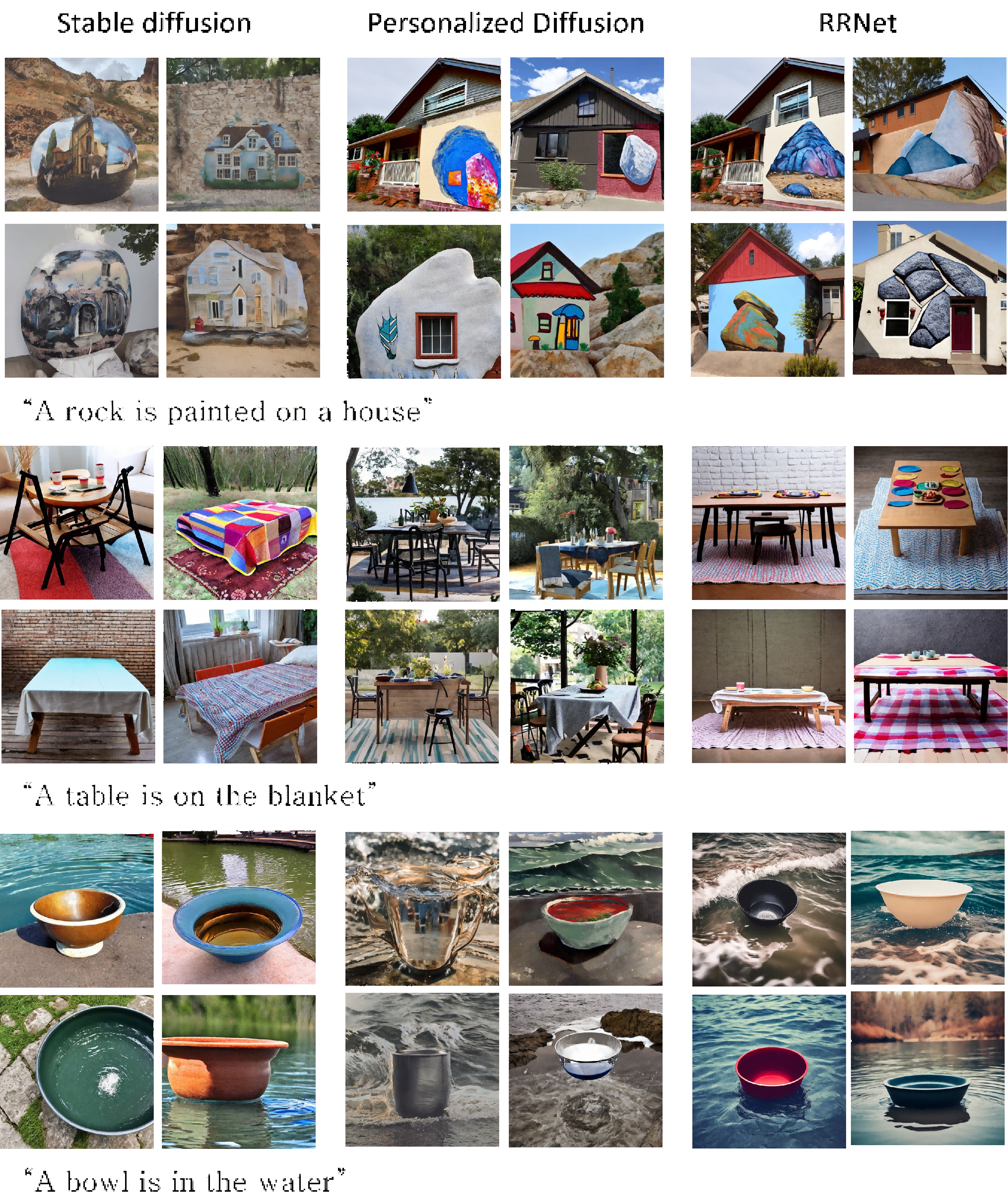}

   \caption{\textbf{Qualitative Comparisons.} Our method outperforms the baselines in terms of the relationship generation.}
   \label{fig:compare}
\end{figure}

\subsection{Comparisons}\label{Comparisons}
\begin{table*}[!h]
  \centering
    \begin{tabular*}{1\linewidth}{@{}lcccccc@{}}
    \toprule
    Method & Position(Qwen) $\uparrow$ & Position(LLaVA)$\uparrow$ &  Action(Qwen) $\uparrow$ & Action(LLaVA)$\uparrow$ & OGA$\uparrow$& FID$\downarrow$\\
    \midrule
    RR Dataset &0.849 & 0.763 &0.616 & 0.652 & 1.000& N/A\\
    Stable Diffusion &0.467 & 0.542 &0.399 & 0.543 &0.898&83.73\\
    Personalized Diffusion & 0.509& 0.558 & 0.273& 0.518 &0.844&91.93\\
    RRNet ($\lambda$=0.2) & 0.564&0.597 & 0.469&0.565&0.937&89.13\\
    RRNet ($\lambda$=0.4) & 0.646&0.648& 0.492&0.603 &0.964&94.73\\
    RRNet ($\lambda$=0.6) & 0.697 & 0.684 & \textbf{0.500} & 0.632 &0.970&100.78\\
    RRNet ($\lambda$=1) & \textbf{0.729}& \textbf{0.724}& 0.490&\textbf{0.651} & 0.971 &110.01\\
    \bottomrule
  \end{tabular*}
  \caption{\textbf{Quantitative Results of Positional Relation Generation.} We evaluate the generation accuracy through Qwen-VL-Chat and LLaVA chatbots, and the generated image quality is evaluated via FID. The Position and Action represent for evaluation on positional relationships part and action relationships part of the dataset, respectively. Object generation accuracy (OGA) is the average calculated from all position and action relations.}
  \label{tab:Quantitative}
\end{table*}

\noindent\textbf{Quantitative analysis.} Table \ref{tab:Quantitative} presents the quantitative results on positional and action relationships generation. Our method obtains a better relationship generation accuracy and object generation accuracy compared to baselines. 
By increasing the adjustment weight $\lambda$ in equation \ref{eq:adjust}, the RRNet can achieve a higher generation accuracy. 

However, it can be observed that a larger $\lambda$ could also leads to a larger FID score, indicating lower image fidelity. To balance relationship generation accuracy and image diversity, we discovered that setting $\lambda$ between 0.4 and 0.6 significantly improves relationship accuracy while preserving image quality.

%


\noindent\textbf{Qualitative analysis.} 
Figure \ref{fig:change} and Figure \ref{fig:cover}(a) illustrate the interpretability of our adjustment vector approach. As $\lambda$ increases, the generated images progressively shift to the correct relationship direction. For example, the second row of Figure \ref{fig:change} shows an astronaut transitioning from riding a horse ($\lambda=0$) to eventually leading the horse ($\lambda=0.6$). This progression visually illustrates the transition of relationships towards the correct direction, demonstrating RRNet's capability in rectifying relations.

In Figure \ref{fig:compare}, we compare our method against baselines. Original SD generations often mix relationship directions, failing to distinguish them accurately.  The personalized diffusion approach, Personalized diffusion, while learning the relationship, often alters object semantics. As observed in the bottom row,  it successfully learns the meaning of ``\texttt{In water}'', but sacrifices the meaning of ``\texttt{A bowl}'', which in turn incorrectly generates cup and bucket similar to bowl. Our approach, in contrast, correctly portrays relationship directions and maintains the objects' original meanings in the generated images.




\subsection{User Study}
\noindent\textbf{Setup.} We conducted a user study with 63 evaluators to evaluate relationship generation accuracy. Ten relationships were chosen from our dataset, with images generated by RRNet and baselines. For each, evaluators selected the image that best depicted the described relationship from two randomly chosen images per method.

\noindent\textbf{Results.} The outcomes of this user study are detailed in Table~\ref{tab:User_Study}. We observe that the generation of RRNet are more favored by evaluators, with an absolute improvement of 68.1\% on personalized diffusion and 58.58\% on SD.

\begin{table*}[!htb]
  \centering
    \begin{tabular*}{\linewidth}{@{}lccccc@{}}
    \toprule
    Method & Position(Qwen) $\uparrow$ & Position(LLaVA)$\uparrow$ &  Action(Qwen) $\uparrow$ & Action(LLaVA)$\uparrow$ & OGA$\uparrow$\\
    \midrule
    RRNet & \textbf{0.697} & \textbf{0.684} & \textbf{0.500} &\textbf{ 0.632} &\textbf{0.970}\\
    w/o HGCN  & 0.509& 0.558 & 0.273& 0.518 &0.844 \\
    w/o negative loss & 0.534 & 0.633 & 0.451 & 0.555 &0.949\\
    w/o node disentanglement & 0.543 & 0.635 & 0.426 & 0.535 &0.938\\
    
    \bottomrule
  \end{tabular*}
  \caption{\textbf{Quantitative Results of Variants of RRNet.}}
  \label{tab:ablation}
\end{table*}
\begin{table}
  \centering
    \begin{tabular}{@{}lc@{}}
    \toprule
    Method & User Preference Rate $\uparrow$ \\
    \midrule
    Stable Diffusion & 16.98\% \\
    Personalized Diffusion & 7.46\% \\
    RRNet ($\lambda$=0.6) & \textbf{75.56\%} \\
    \bottomrule
  \end{tabular}
  \caption{\textbf{Results of User Study.} The result shows the percentage of evaluators prefer the image produced by RRNet verse baselines.}
  \label{tab:User_Study}
\end{table}

\subsection{Generalization to Unseen Objects}
\begin{figure}[!htb]
  \centering
   \includegraphics[width=\linewidth]{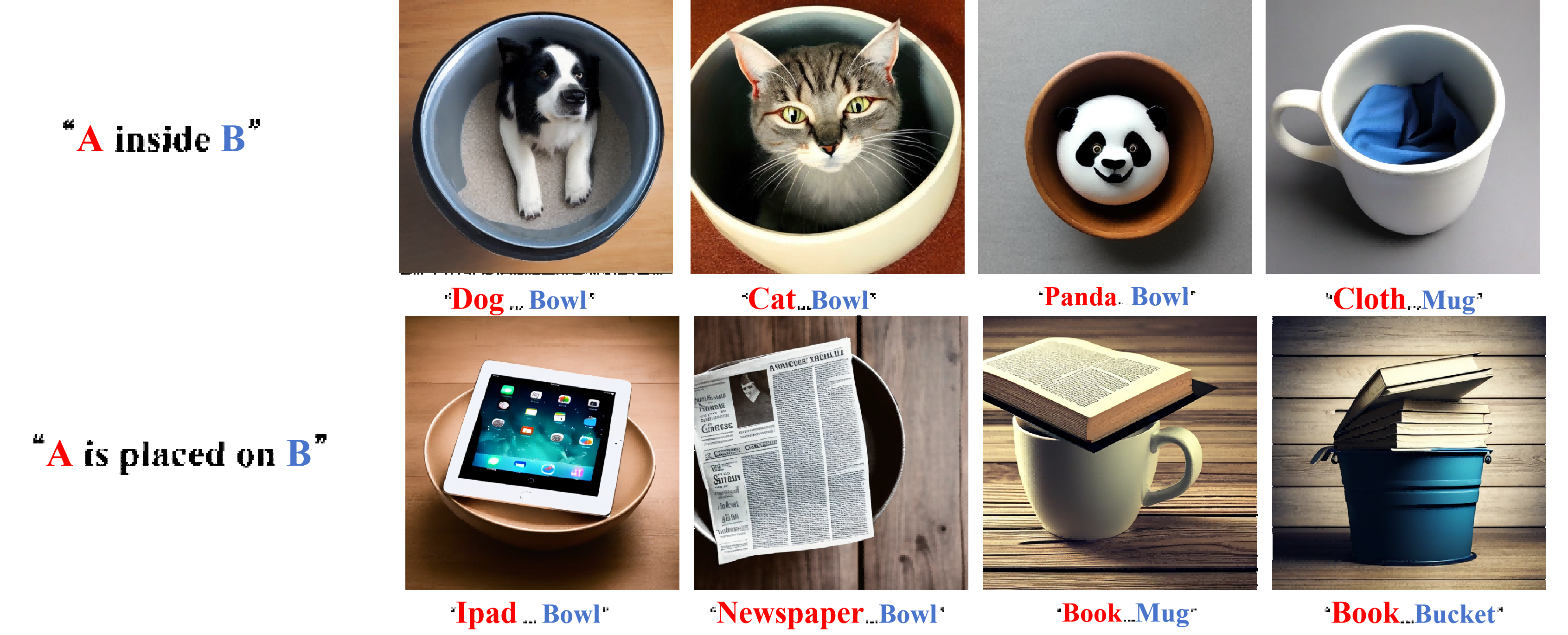}
   \caption{\textbf{Example of Generalization.} By switching the objects in the prompts, RRNet can still generate correct relationships.}
   \label{fig:gene}
\end{figure}
In our framework, for each set of OSPs, the RRNet is trained with images contains only one pair of objects. It is normal to wonder if the trained RRNet can generalize well to the unseen objects.

As shown in Figure \ref{fig:gene} and Figure \ref{fig:cover}(b), by constructing new graphs, RRNet demonstrates its ability to handle many objects unseen in the training dataset. 


\subsection{Ablation Study}
\begin{figure}[!htb]
  \centering
   \includegraphics[width=1\linewidth]{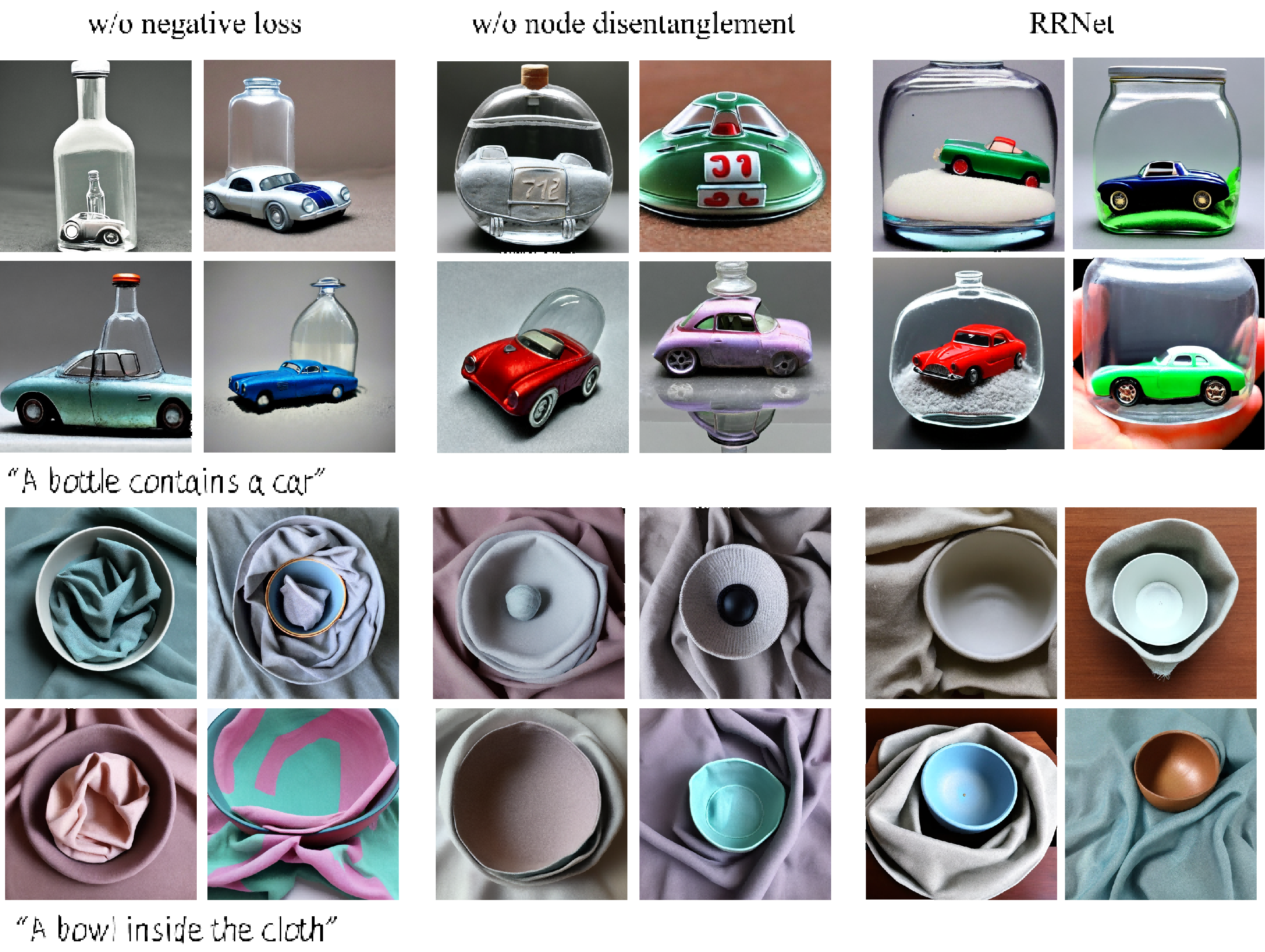}
   \caption{\textbf{Ablation Results.}}
   \label{fig:ablation}
\end{figure}

Here, we demonstrate the importance of our three designs, 1) HGCN 2) negative loss, and 3) object node disentanglement. For a fair comparison, we kept the training hyperparameters the same and set $\lambda$=0.6 during image generation.

\noindent\textbf{HGCN.} Excluding the HGCN, which accounts for graph structure, RRNet reverts to a basic personalized diffusion model. This scenario has been analyzed and compared in Table~\ref{fig:compare}. The results indicate that the construction of relational graphs is crucial for effective relation modeling in RRNet.

\noindent\textbf{Negative loss.} As depicted in Figure \ref{fig:ablation}, the absence of negative loss leads to generated images that somewhat conform to both relational directions. For instance, in the upper row, images of ``\texttt{A bottle contains a car, and a car contains a bottle}'' are generated. Meanwhile, as can be seen from Table \ref{tab:ablation}, without negative loss, the performance of the model decreases a lot.



\noindent\textbf{Object node disentanglement.} Omitting this component leads to the blending of semantics of nodes representing different objects. As shown in Figure \ref{fig:ablation}, this results in generated images with merged features, such as a bottle with car-like traits or a bowl resembling cloth. Quantitative results in Table \ref{tab:ablation} further show reduced accuracy in generating both relationships and objects.

\section{Conclusion}
We presented a novel framework, named RRNet, to enhance the ability of diffusion models in generating images with more accurate relation directions. The key of our approach is to employ a HGCN to explicitly model the relational direction present in the prompts. 
RRNet will produce an adjustment vector to rectify the direction of the relations in the text embeddings that are provided to SD. 
We experimentally demonstrate the interpretability as well as the robust performance of our approach.

\section*{Acknowledgements}
This research is supported by the Ministry of Education, Singapore, under its Academic Research Fund Tier 2 (Award Number: MOE-T2EP20122-0006).

{\small
\bibliographystyle{ieee_fullname}
\bibliography{egbib}
}

\clearpage
\appendix

This supplementary document offers additional comparisons and detailed information regarding our proposed RRNet model. In Sec \ref{AC}, we offer a further comparison between proposed RRNet and more baseline methods. In Sec \ref{RV}, we present more visualized results. We then provide additional analysis of $V_{eot}$ in Sec \ref{AA}. The implementation and additional experimental details are included in Sec \ref{ES}. Finally, we discuss the limitations of RRNet in Sec \ref{L}.
\section{Additional Comparisons}\label{AC}
\begin{table}[!h]
  \centering
    \begin{tabular}{@{}lc@{}}
    \toprule
    Method & GPU Memory Usage (MiB)\\
    \midrule
    TI & 25458 \\
    Unet Update & 30060 \\
    
    Unet Update + TI & OOM \\
    RRNet & 25040 \\
    \bottomrule
  \end{tabular}
  \caption{\textbf{Memory usage of training.}}
  \label{tab:memory}
\end{table}
\begin{table*}[!h]
  \centering
    \begin{tabular*}{1\linewidth}{@{}lcccccc@{}}
    \toprule
    Method & Position(Qwen) $\uparrow$ & Position(LLaVA)$\uparrow$ &  Action(Qwen) $\uparrow$ & Action(LLaVA)$\uparrow$ & OGA$\uparrow$& FID$\downarrow$\\
    \midrule
    TI  & 0.560 & 0.592 & 0.443 & 0.574 &0.929&91.634\\
    Unet Update  & 0.473 & 0.563 & 0.404 & 0.529 &0.912&87.294\\
    RRNet ($\lambda$=0.2) & 0.564&0.597 & 0.469&0.565&0.937&89.13\\
    RRNet ($\lambda$=0.6) & \textbf{0.697} & \textbf{0.684} & \textbf{0.500} & \textbf{0.632} &\textbf{0.970}&100.78\\
    \bottomrule
  \end{tabular*}
  \caption{\textbf{Further comparisons.} }
  \label{tab:further comparison}
\end{table*}

\noindent\textbf{Additional baselines.} To provide a more comprehensive comparison, we compare with three other baseline methods. 

\begin{figure}[!h]
  \centering
   \includegraphics[width=1\linewidth]{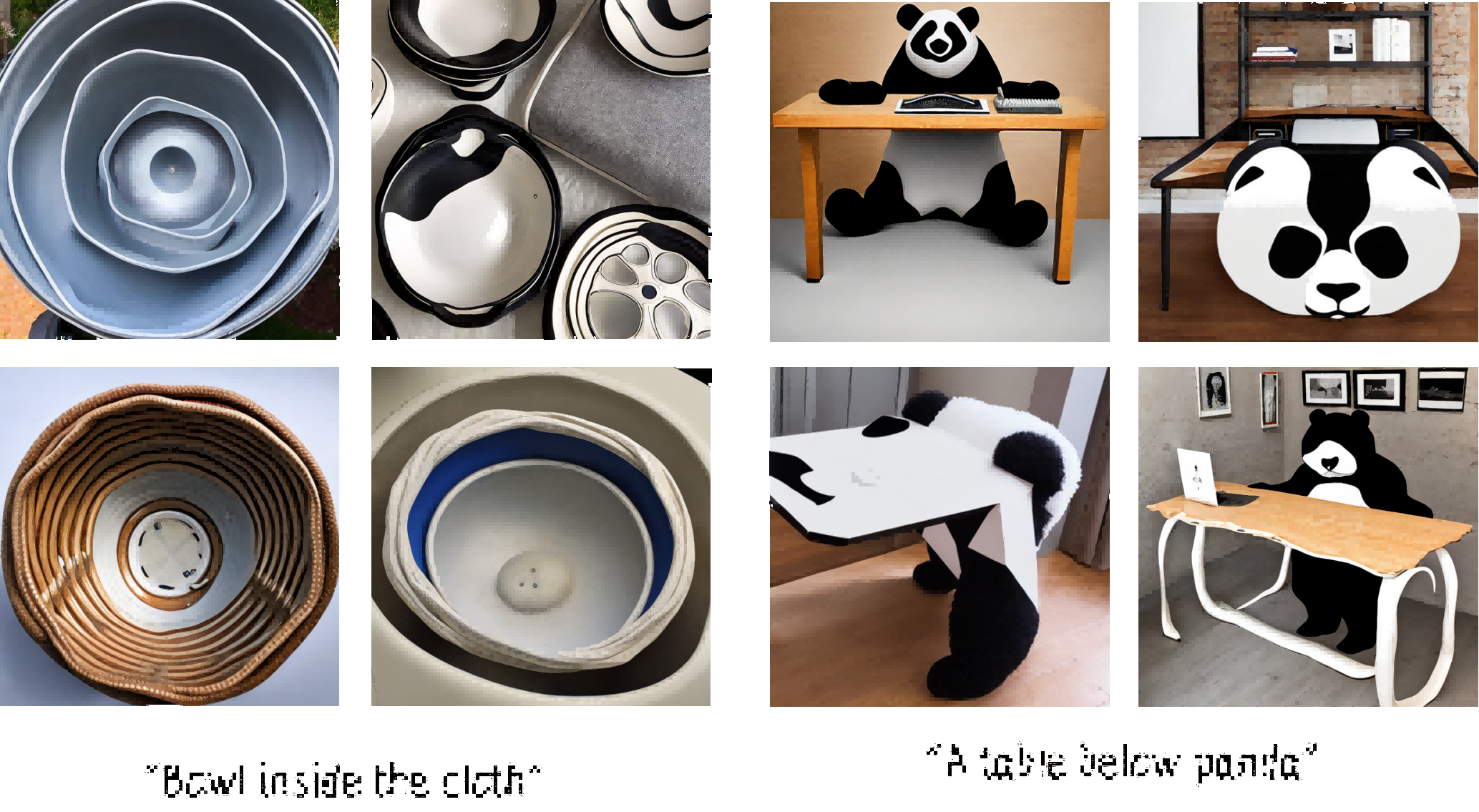}

   \caption{In our dataset, the Reversion fails to achieve accurate relationship generation.}
   \label{fig:reversion}
\end{figure}

\begin{itemize}
    \item \textbf{Textual Inversion.} The first method is a variation of textual inversion \cite{Gal_Alaluf_Atzmon_Patashnik_Bermano_Chechik_Cohen}. In this approach, we employ the its loss function to fit the diffusion model to the given set of images by  optimizing the input embeddings. 
    \item \textbf{Unet Update.} The second method involves maintaining the text encoder of SD fixed while fine-tuning the Unet component. Specifically, we implement Low-Rank Adaptation (LoRA) \cite{Hu_Shen_Wallis_Allen-Zhu_Li_Wang_Chen_2021} to prevent catastrophic forgetting in SD. The training parameters are set with a learning rate of 1e-4 and a rank of 4, over a duration of 100 epochs. 
    \item  \textbf{Reversion.} Furthermore, we tested the Reversion method on our dataset. The empirical results indicated that the images generated through Reversion were of poor quality. This suboptimal performance could be attributed to the inherent complexities associated with the Relation Rectification Task. Due to these findings, we decided against including the quantitative results of this method in our report. For reference, examples of these generated images are depicted in Figure \ref{fig:reversion}. 
\end{itemize}


A potential fourth method, optimizing the text encoder and Unet parameters simultaneously, was considered but deemed impractical due to its excessive computational demands. The memory requirements for training using different methods are compared in Table \ref{tab:memory}. Training with a batch size of 1 is not unfeasible on a single V100-32GB. 


\noindent\textbf{Quantitative analysis.} The results are presented in the Table \ref{tab:further comparison}. 
It can be observed that among the baselines, TI demonstrates higher performance in relationship generation accuracy metrics compared to Unet Update.
Our RRNet, employing a mere $\lambda=0.2$, closely approaches the performance of TI across various relationship generation accuracy metrics, while also maintaining a lower FID.
Increasing $\lambda$ to 0.6 significantly elevates RRNet's performance above the baselines in all relationship generation accuracy metrics. Specifically, in the Position (Qwen) and Action (Qwen) metrics, RRNet ($\lambda$=0.6) outperforms TI by 13.7\% and 5.7\%, respectively. Furthermore, when using LLaVA as the detector, RRNet ($\lambda$=0.6) also exceeds TI by 9.2\% and 5.8\% respectively in position and action relationship generation metrics. In terms of OGA, RRNet ($\lambda$=0.6) surpasses TI by a margin of 4.1\%.

For those two baselines, since the issue with the text encoder treating sentences as a Bag of Words (BOW) remains unresolved, they still struggle to accurately distinguish the direction of relationships. They tend more to directly fit the dataset rather than truly learning to represent the relationships between objects. 

\section{Results Visualization}\label{RV}
\begin{figure*}[!h]
  \centering
   \includegraphics[width=1\linewidth]{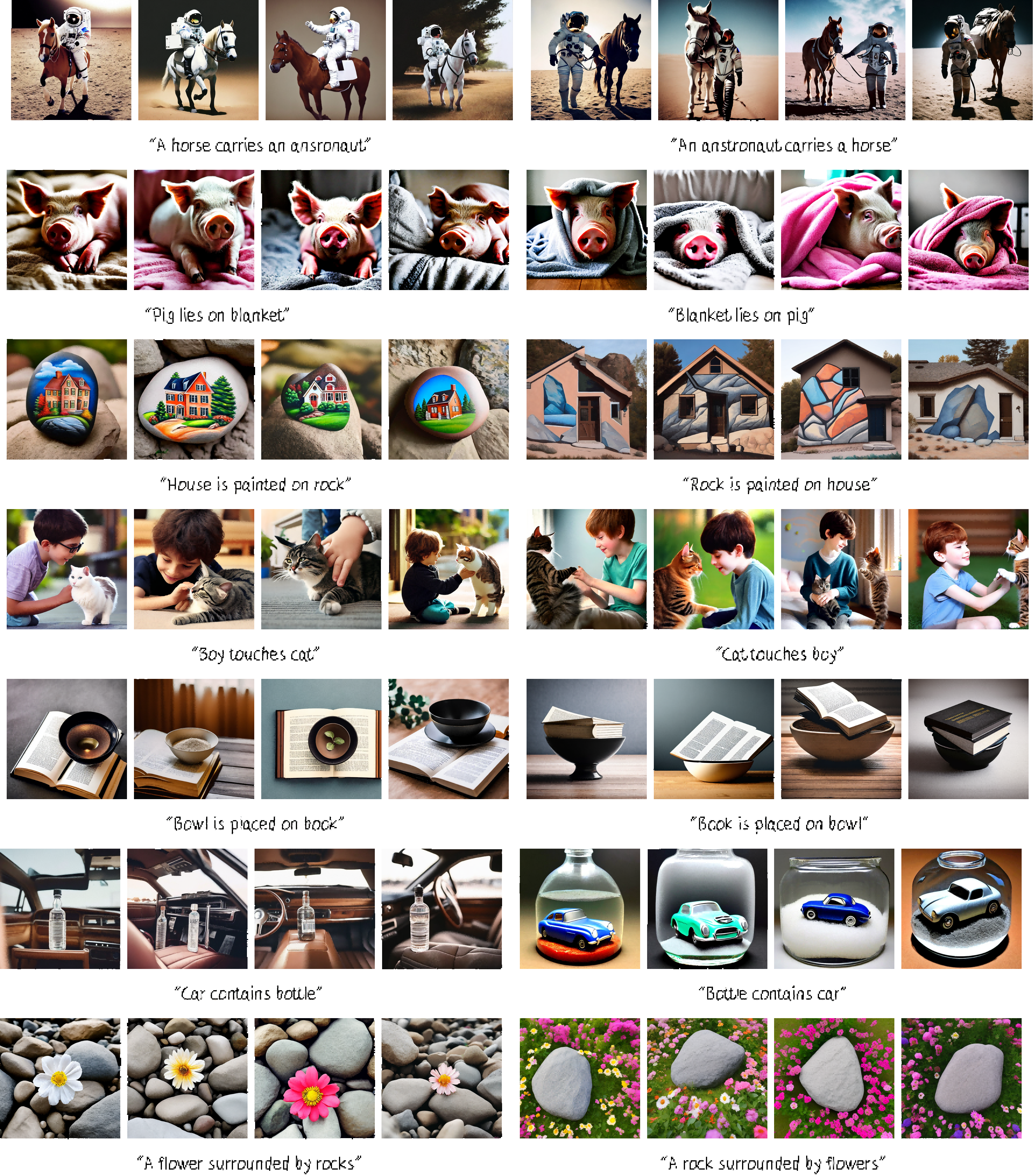}

   \caption{Additional results of action relation rectification with RRNet.}
   \label{fig:visual_verb}
\end{figure*}

\begin{figure*}[!h]
  \centering
   \includegraphics[width=1\linewidth]{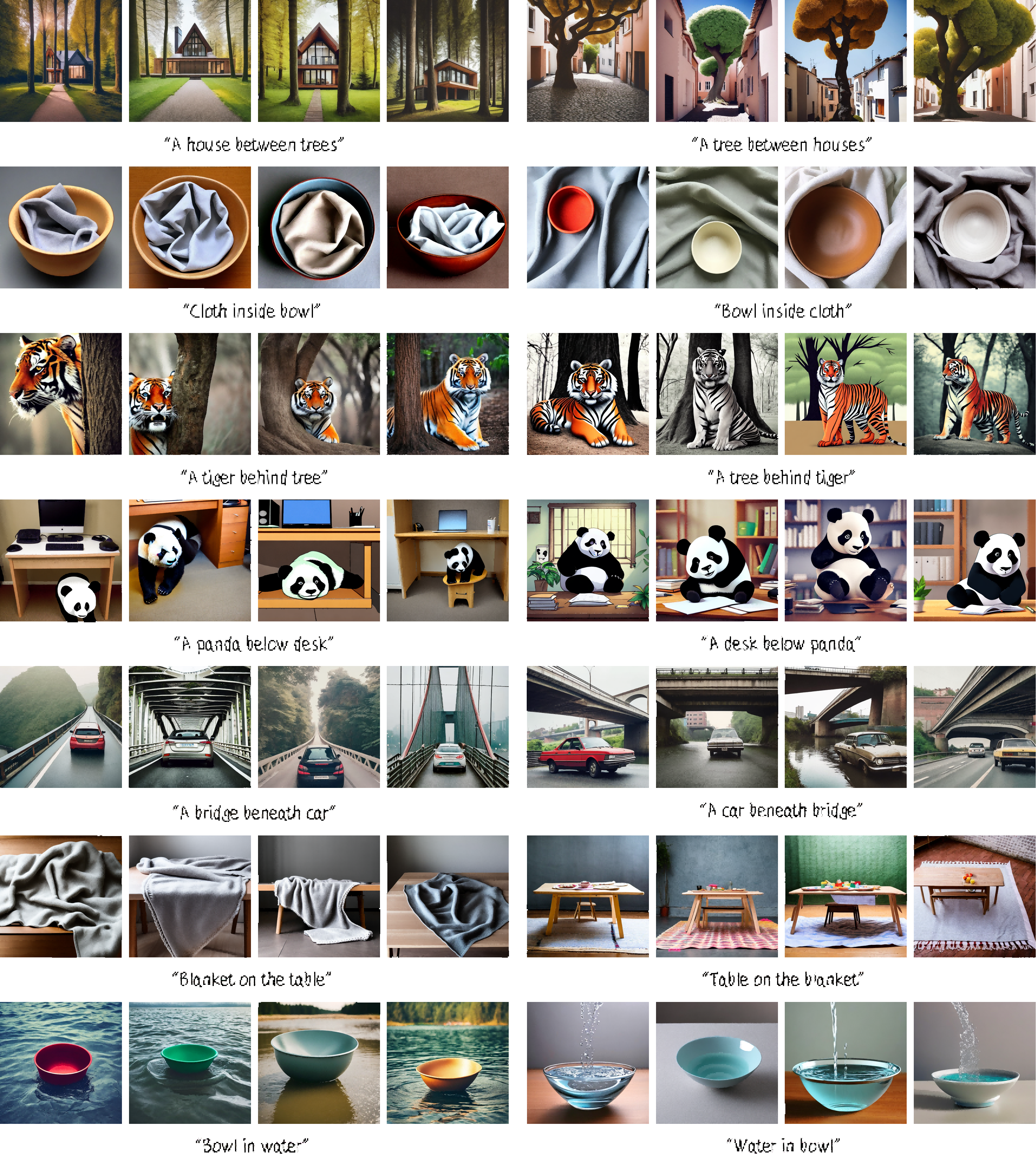}

   \caption{Additional results of positional relation rectification with RRNet.}
   \label{fig:visual_noun}
\end{figure*}

We provide additional qualitative results to demonstrate the effect of RRNet on relation rectification for SD. The generation results of action OSPs are shown in Figure \ref{fig:visual_verb} and the results of positional OSPs are showcased in Figure \ref{fig:visual_noun}.

It can be observed that with RRNet's assistance, SD can accurately generate relationships in both directions.

\begin{figure}[!h]
  \centering
   \includegraphics[width=1\linewidth]{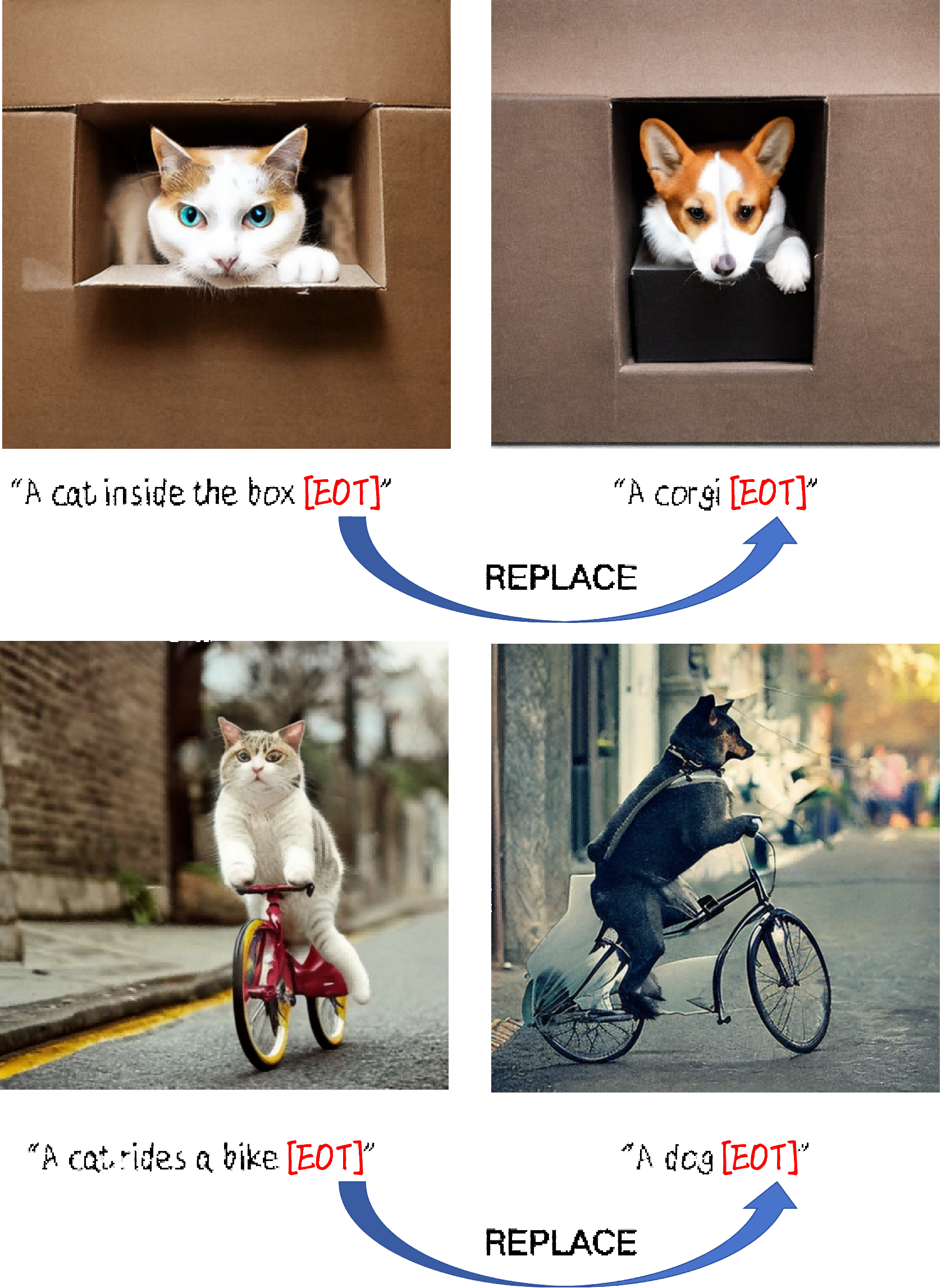}

   \caption{After replacing the $V_{eot}$, a phenomenon of entity replacement occurred.}
   \label{fig:replace}
\end{figure}

\section{Additional Analysis of $V_{eot}$ }\label{AA}


In the Sec \ref{approach}, we identify the $V_{eot}$ as pivotal in generating relationships. Specifically, as detailed in previous work \cite{Chen_Laina_Vedaldi_2023}, the $V_{eot}$ plays a crucial role in foreground generation, encompassing both the objects and their interrelationships.

Here, we conducted an experiment by replacing the $V_{eot}$ of two prompts to validate its significance.

\noindent\textbf{Observation.} Our experiment, illustrated in the first row of Figure~\ref{fig:replace}, involved replacing the $V_{eot}$ from ``\texttt{A corgi}'' with that from ``\texttt{A cat inside the box}''. This resulted in an image of  ``\texttt{A corgi inside the box}''. This experiment suggests that the primary role of $V_{eot}$ is to control the foreground layout, dictating where each object appears. Since the semantic of ``\texttt{A corgi}'' closely match that of ``\texttt{A cat}'', it would match the anchor for ``\texttt{A cat}'' and instead generate a corgi.

\noindent\textbf{Discussion.} We uncovered a fascinating aspect of $V_{eot}$ related to the diffusion generation mechanism. Analogous to painting, the $V_{eot}$ shapes the basic layout of the foreground, setting anchor points for each object's generation. It establishes an anchor point for each foreground object to be generated, without detailing each object specifically.  The word embeddings of these objects then align with the closest anchor, leading to their manifestation at specific locations. The occurrences in Figure [Mask Phenomenon] can be seen as a failure in object generation due to the lack of precise positional information in $V_{eot}$.


Therefore, in our work, we achieve the effect of correctly controlling the generation of the foreground by adjusting $V_{eot}$.





\section{Experiment Settings}\label{ES}
{\subsection{Detailed dataset statistics.} 
The RR dataset comprises 21 relationships, consisting of 8 positional and 13 action types. Each relationship includes 4 types of prompts, two for OSPs represented as $<A, R, B>$ and $<B, R, A>$, and two generated from template sentence ``\texttt{This is a photo of \{obj\}}'' for object disentanglement purposes.
We collect 3-5 images for each prompt to serve as exemplars for training.}

\subsection{Relationship Detection with Chatbots.} 
\noindent\textbf{Prompts for relationship detection.} We employ Vision-language chatbots to facilitate the detection of relationships in images.
To ensure these chatbots focus more on the relationships between objects, we have developed a series of prompt templates. For a sentence can be abstracted as triplet $<A,R,B>$, prompt templates are listed follows:
\begin{enumerate}
 \item ``\texttt{Is there any {object A} in the image?}''
 \item ``\texttt{Is there any {object B} in the image?}''
 \item ``\texttt{Are both {object A} and {object B} present in the image?}''
 \item ``\texttt{Can you infer the relationship that exists between \textit{object A} and \textit{object} B in the image?}''
 \item ``\texttt{Is there {ARB} or {BRA} or neither?}''
\end{enumerate}

``\texttt{object A}'' and ``\texttt{object B}'' are placeholders substituted by entities in the OSPs. ``\texttt{ARB}''  and ``\texttt{BRA}'' are OSPs.
During the evaluation, each generated image undergoes assessment through above five questions.

\noindent\textbf{Object detection.} Question 1 and Question 2 are designed for object detection. During our experiments, we observed a high false-positive rate in object detection using the first two questions, primarily because the model occasionally generates a composite object embodying features of both objects A and B. Consequently, even if there is a single object displaying both sets of features, the chatbots are likely to affirmatively respond to the first two questions. To mitigate this issue, we introduced Question 3, aimed at filtering out objects that exhibit mixed features. For object detection, a generation is deemed correct only if it successfully passes Questions 1, 2, and 3.

\noindent\textbf{Relationship detection.} Question 4 is designed to steer the chatbot towards creating contexts that emphasize the relationships between objects, thereby setting the stage for Question 5.
Regarding relationship generation, a relationship generation is classified as correct only if the chatbot's response to the Question 5 aligns with the prompt for generating the given image.


\subsection{Additional Implementation details.} \label{AID}

\noindent\textbf{Implementation details of HGCN.} In our implementation, the HGCN is built with DGL \cite{wang2019dgl}. We use the HGCN based on Graph Attention Networks (GAT) \cite{Liu_Zhou_2020}. The dimension of HGCN's hidden layers is 512. The learning rate of HGCN is set to 3e-4. The optimizer used is AdamW \cite{loshchilov2019decoupled}. The training batch size is set to 1.


\section{Limitations}\label{L}
\subsection{Unseen concepts}
\begin{figure}[!h]
  \centering
   \includegraphics[width=1\linewidth]{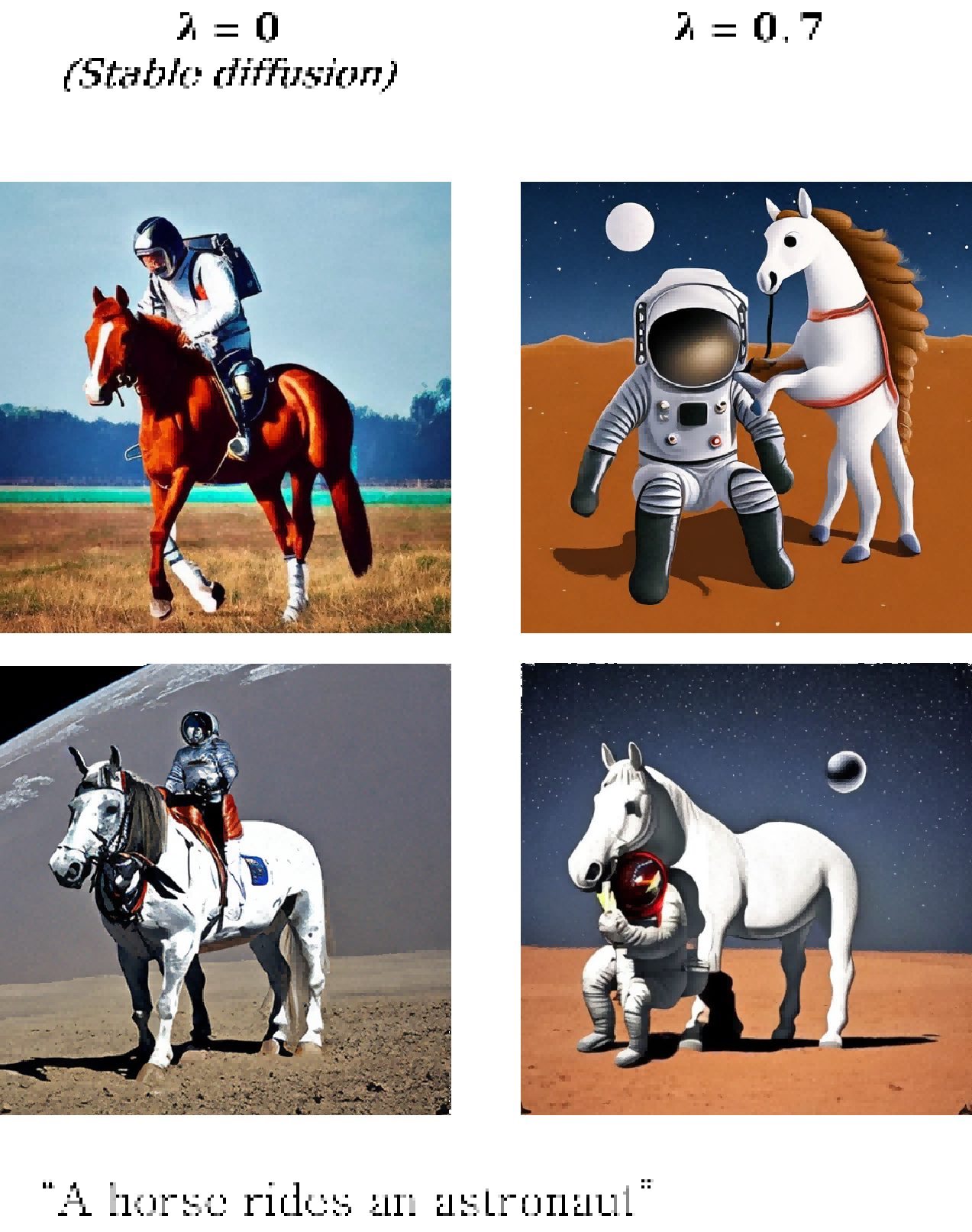}

   \caption{\textbf{Failure case.} The model has no sense of how horse rides on other objects.}
   \label{fig:limitatons}
\end{figure}

\begin{figure}[!h]
  \centering
   \includegraphics[width=1\linewidth]{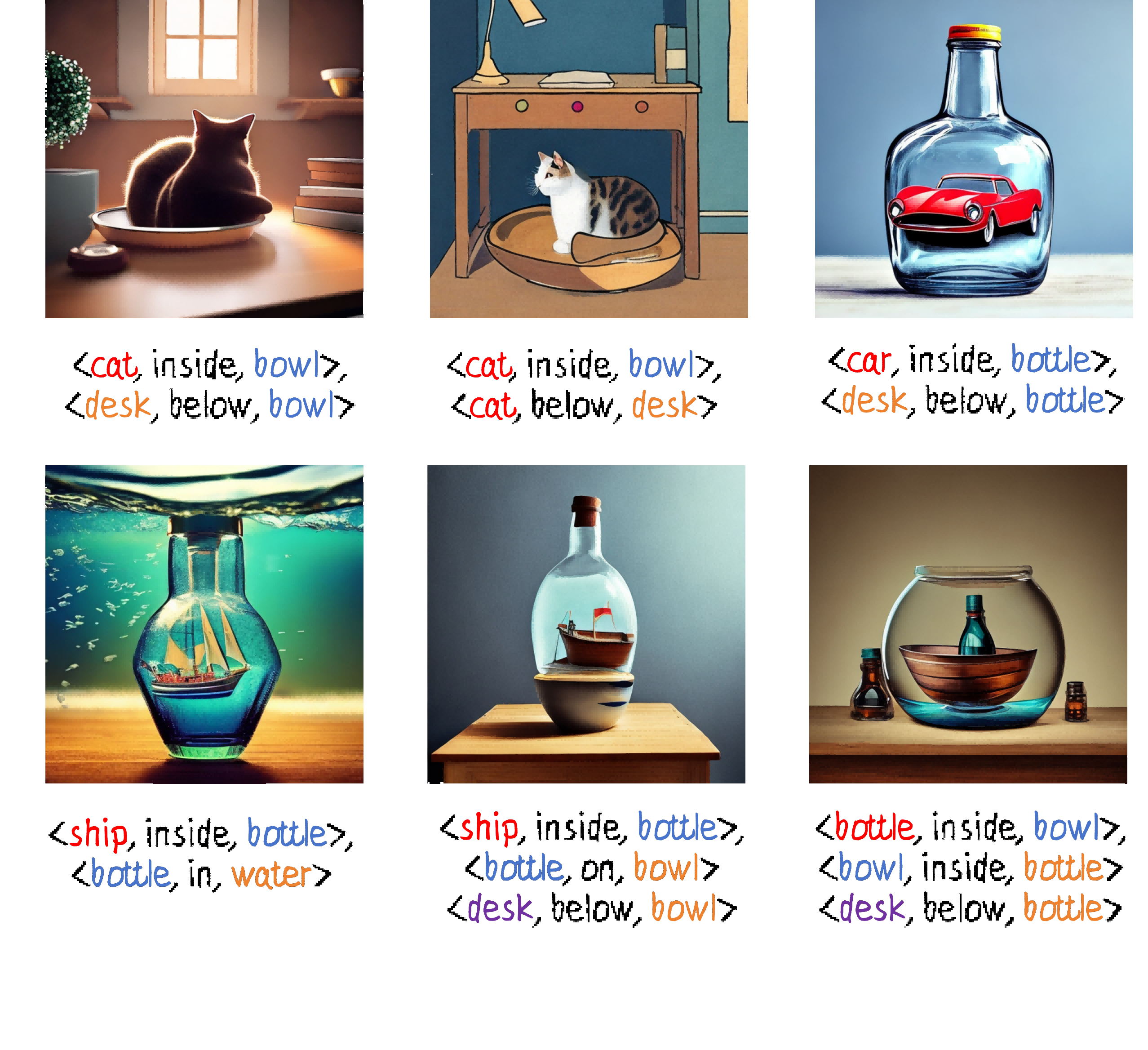}
   \caption{\textbf{Generation of Complex Relationships.}}
   \label{fig:complex rel}
\end{figure}

Our method is capable of steer the generation of SD by adjusting the direction of the relations in the text embeddings. Since we do not modify any parameters in denoising network, there are limitations to our approach for concepts that do not exist in SD.
We show a failure case in Figure \ref{fig:limitatons}. 
Although our method separates the ``\texttt{A horse rides an astronaut}''  from ``\texttt{An astronaut rides a horse}'', SD lacks the concept of a horse riding anything, leaving the RRNet directionless in adjusting the horse-astronaut relationship. Consequently, most generated images are mere variations of existing dataset images. we observe that for such abstract relations, we need a larger $\lambda$ to ensure the generated images meaningful, which in term undermines the images' quality.

\subsection{Multi-relationships generation}

we employed multiple RRNets, initially \emph{trained on simple paired relations}, to handle more complex scenarios in image generation. The results are illustrated in the Figure \ref{fig:complex rel}. Although generated results looks reasonable, there is a noticeable drop in performance as the complexity of relationships and objects increases.
The occurrence of this phenomenon is likely due to multiple RRNETs simultaneously adjusting the $V_{eot}$, resulting in semantic confusion within $V_{eot}$.
Therefore, we believe that exploring how to construct a more complex graph to generate one adjustment vector capable of jointly rectifying multiple relational semantics is a highly promising avenue for future research.




\end{document}